\UseRawInputEncoding
\documentclass{article}

\usepackage{microtype}
\usepackage{graphicx}
\usepackage{subcaption}
\usepackage{booktabs} % for professional tables
\usepackage{pifont} % added by authors

\usepackage[most]{tcolorbox}   % most 또는 breakable 라이브러리 포함
\usepackage{fvextra} 
\usepackage{cuted}      % 전체폭 non-floating strip
\usepackage{tabularx}
\newcolumntype{Y}{>{\raggedright\arraybackslash}X}
\usepackage{hyperref}
\usepackage{fvextra}
\usepackage{xcolor}

\usepackage[accepted]{icml2026}

\usepackage{amsmath}
\usepackage{amssymb}
\usepackage{mathtools}
\usepackage{amsthm}
\usepackage{thmtools} % added by authors
\usepackage{thm-restate} % added by authors

\usepackage{multirow}
\usepackage{adjustbox}
\usepackage{wrapfig}
\usepackage{pgfplots}
\pgfplotsset{compat=1.18}
\usetikzlibrary{arrows.meta}
\definecolor{navy}{HTML}{1E2761}
\definecolor{teal2}{HTML}{1C7293}
\definecolor{mint}{HTML}{129E8A}
\definecolor{goodg}{HTML}{0E7C66}
\definecolor{badr}{HTML}{B04A3A}

\makeatletter
\newcommand\footnoteref[1]{\protected@xdef\@thefnmark{\ref{#1}}\@footnotemark}
\makeatother

\usepackage[capitalize]{cleveref}
 \AtBeginDocument{%
    \crefname{figure}{Figure}{Figures}%
}
\newcommand{\eqlabelleft}{(}
\newcommand{\eqlabelright}{)}

\creflabelformat{equation}{\eqlabelleft#2#1#3\eqlabelright}

\renewcommand{\algorithmicrequire}{\textbf{Input:}}
\renewcommand{\algorithmicensure}{\textbf{Output:}}

\newcommand{\cmark}{\ding{51}}  % Check mark
\newcommand{\xmark}{\ding{55}}  % Cross mark

\theoremstyle{plain}

\theoremstyle{definition}

\theoremstyle{remark}

\usepackage[textsize=tiny]{todonotes}

\usepackage{amsmath,amsfonts,bm}

\def\eqref#1{equation~\ref{#1}}
\def\1{\bm{1}}

\DeclareMathAlphabet{\mathsfit}{\encodingdefault}{\sfdefault}{m}{sl}
\SetMathAlphabet{\mathsfit}{bold}{\encodingdefault}{\sfdefault}{bx}{n}

\icmltitlerunning{Two-Axis Reasoning for Visually-Grounded Physics Problem Solving}

\usepackage{enumitem}
\begin{document}

\twocolumn[
\icmltitle{Multi-Agent Debate and Visual Information Extraction for SeePhys Pro: A 1st-Place Technical Report from ICML 2026 AI4Math Track 3 Challenge}



  \icmlsetsymbol{equal}{*}

  \begin{icmlauthorlist}
    \icmlauthor{Jiseok Kwak}{kaist}
    \icmlauthor{Suhyeon Jo}{kaist}
    \icmlauthor{Taewoo Kim}{kaist}
    \icmlauthor{Yeongmin Kim}{kaist}
    \icmlauthor{Byeonghu Na}{kaist}
    \icmlauthor{Il-Chul Moon}{kaist,summary}
  \end{icmlauthorlist}

  \icmlaffiliation{kaist}{KAIST}
  \icmlaffiliation{summary}{summary.ai}

  \icmlcorrespondingauthor{Il-Chul Moon}{icmoon@kaist.ac.kr}
  \icmlcorrespondingauthor{Jiseok Kwak}{jskwak@kaist.ac.kr}

  \icmlkeywords{Machine Learning, ICML}

  \vskip 0.3in
]



\printAffiliationsAndNotice{}  % no special notice (required even if empty)

\begin{abstract}
  This technical report presents our approach to Challenge Track~3: SeePhys Pro at the 3rd AI for Math Workshop, where the task is to answer college-level physics questions whose statement and figure may be given partly or entirely as an image. Visual physics problems become substantially harder for large language models when the decisive information resides in a figure rather than in the text, and this modality gap widens as more of the problem migrates into the image. We address the task with a two-stage framework: a visual information extraction stage that re-expresses figure content as solver-readable text to close the modality gap, and a reasoning stage that orchestrates three heterogeneous solvers through multi-agent debate. Our analysis yields two findings: the gain from orchestration comes from reliable answer selection rather than from additional debate, and the value of a figure aid scales with how much of the problem is locked inside the image. The resulting pipeline improves overall accuracy over a single-agent baseline from 0.643 to 0.802 on the public split, and won 1st place on both the public and the private leaderboard (private overall 0.743).
\end{abstract}

\section{Introduction}
\label{sec:intro}

Coordinating multiple large language models has emerged as a scaling axis in its
own right, separate from enlarging any single model's weights. Multi-agent debate
with memory masking (MAD-M$^2$)~\cite{tian2026multiagent} and learned orchestrators
such as Sakana's Fugu~\cite{sakana2026fugu} both surpass the strongest single model
in their pool by coordinating heterogeneous
solvers~\cite{du2024improving,tian2026multiagent,sakana2026fugu}: the coordination
protocol, not the base weights, becomes the lever. We bring this lens to visual
physics problem solving.

SeePhys Pro~\cite{xiang2026seephyspro}, the benchmark for Challenge Track~3 at the
3rd AI for Math Workshop (ICML~2026), is unusually suited to the question. Its
Levels~1--4 re-encode the \emph{same} underlying physics problems at escalating visual
difficulty, from text-only (L1) to image-only (L4), where the entire problem lives
in a single image, so any accuracy difference across levels is attributable to
modality alone. This turns the multimodal--text gap that the ICML~2025 SeePhys
winners observed~\cite{liang2025multimodal} into a controlled measurement, and lets
us study a two-stage pipeline, visual information extraction followed by reasoning,
along the two orthogonal axes the design separates: orchestrating \emph{reasoning}, and closing the \emph{modality
gap}.

Our two findings mirror these axes. On text-only L1, where no modality gap exists,
orchestration's payoff is reliable answer \emph{selection} rather than additional
reasoning: a cheap value-aware canonicalizer matches a premium adjudicator in the tie-breaking role;
once value-equal ties are resolved, the premium adjudicator adds no correct answers. On L2--L4, holding the
reasoning stack fixed and varying only the visual front-end, the value of a figure
aid scales with the size of the modality gap: large on image-only L4, where visual
extraction is the bottleneck, and negligible where the figure merely duplicates the
text. We extend caption-assisted reasoning~\cite{liang2025multimodal} from prose
captions to a structured SVG representation that preserves figure geometry.

Our contributions are as follows:
\begin{itemize}[topsep=2pt,itemsep=3pt,parsep=0pt,leftmargin=1.4em]
  \item \textbf{Multi-agent debate for physics problem solving.} We bring
  multi-agent debate to physics, running three frontier solvers drawn from
  different model families, and show that together they substantially outperform
  a single-solver baseline. The improvement comes from reliably selecting among
  the answers the solvers produce, rather than from the debate itself.
  \item \textbf{Visual information extraction that closes the modality gap.} We
  re-express each figure as text the solvers can read: a prose transcription, a
  structured SVG, and, at the hardest level, a crop-and-OCR front-end. The benefit
  of this extraction grows in step with how much of the problem is locked inside
  the image.
  \item \textbf{First place on both leaderboards.} Combining the two, our system
  improves on a single-pass baseline at every level of SeePhys Pro and finished
  first on both the public and the private leaderboard of Challenge Track~3.
\end{itemize}

\section{Preliminary}
\label{sec:prelim}

\subsection{Multi-Agent Debate with Memory Masking (MAD-M\texorpdfstring{$^2$}{2})}
\label{subsec:prelim_madmm}
Multi-agent debate coordinates several LLMs that independently propose answers,
critique one another, and revise over successive rounds, aggregating by majority
vote; it is a canonical instance of orchestration as a scaling
lever~\cite{du2024improving}. Its weakness is that unreliable intermediate reasoning
is shared verbatim between rounds, so a confident but wrong chain can propagate and
contaminate the other agents. MAD-M$^2$ adds an explicit memory-masking step: at the
start of each round it evaluates every prior-round solution and masks the unreliable
ones, so agents re-solve from cleaner context rather than compounding earlier
errors~\cite{tian2026multiagent}. On mathematical and logical reasoning benchmarks
this yields a clear progression over plain chain-of-thought and vanilla debate (e.g.,
GSM8K with Qwen2.5-7B: CoT $64.8 \rightarrow$ MAD $81.2 \rightarrow$ MAD-M$^2$
$89.0$), showing that the coordination protocol drives the gain. Our reasoning stage
(\cref{subsec:method_reasoning}) instantiates MAD-M$^2$ over three \emph{heterogeneous}
frontier solvers and departs from pure majority voting by adding a value-aware
Canonicalizer.

\subsection{From Figures to Structured Representations via LLM Captioning}
\label{subsec:prelim_caption}
The same model often solves a physics problem more reliably from text than from an
equivalent figure, motivating \emph{caption-assisted reasoning}: re-expressing a
figure as text the model can read, so that visual content enters the reasoning
channel the model handles best. The ICML~2025 SeePhys 1st-place solution framed this
as bridging the multimodal--text gap with model-generated prose
captions~\cite{liang2025multimodal}. Prose captions, however, discard the geometric
and relational structure that many physics figures encode: spatial layout, region
boundaries, connectivity, and part--whole relations that are often decisive for the
correct analysis. We therefore extend this lineage from flat prose to a
\emph{structured} representation~\citep{lu2021intergps,zhang2022pgdp,rodriguez2023starvector}: a figure-to-SVG rendering that preserves geometry
and labels, paired with a prose transcription and, at the highest visual difficulty,
a crop-and-OCR front-end. This background is generalized by the modality-adaptive
extraction stage (\cref{subsec:method_visual}).

\begin{figure*}[t]
    \centering
    \includegraphics[width=\linewidth]{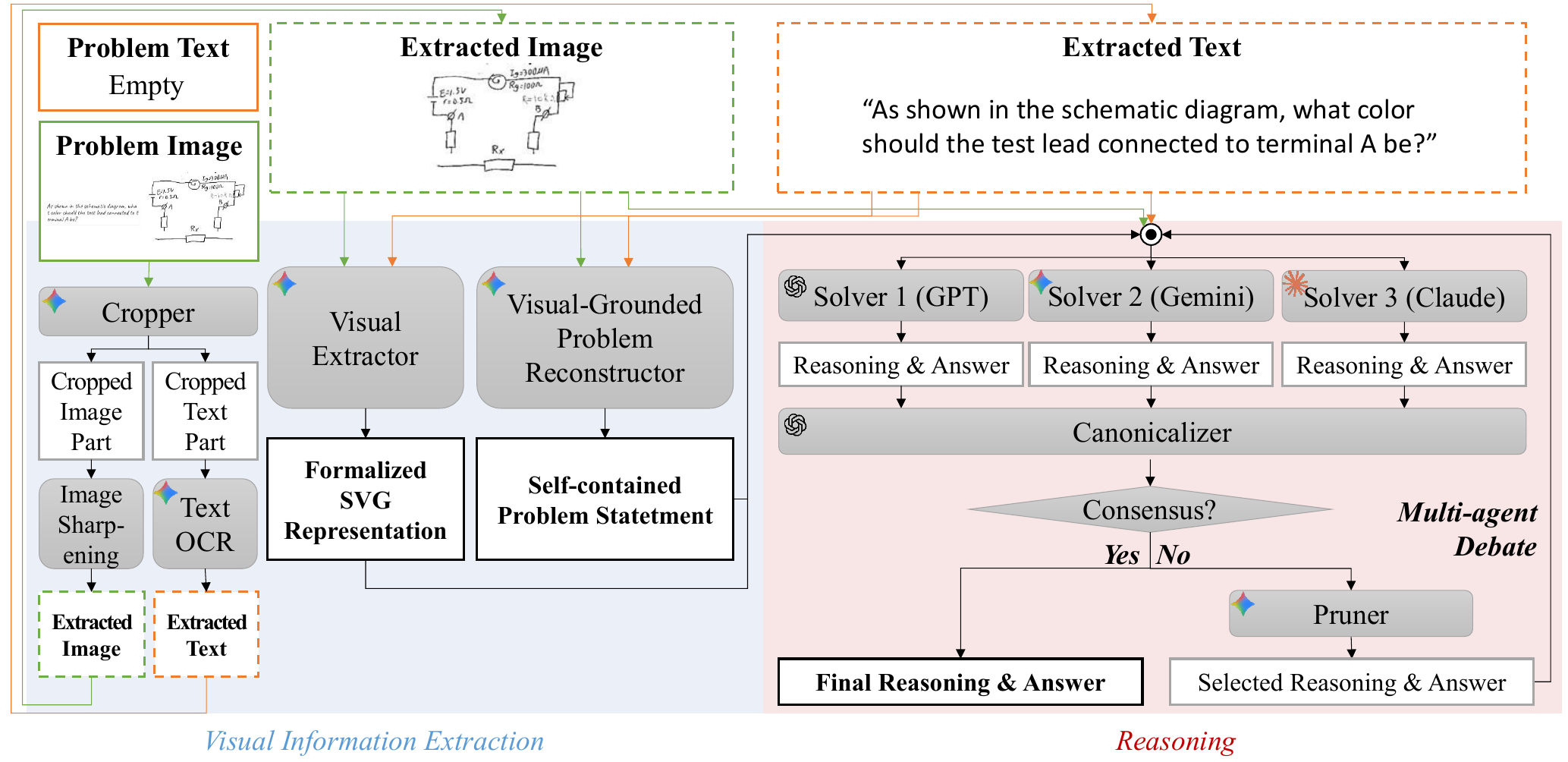}
    \caption{Proposed pipeline for end-to-end visual reasoning on Level 4 problems in the SeePhys Pro dataset.}
    \label{fig:overview_level4}
\end{figure*}

\begin{table*}[tp]
    \centering
    \caption{Level-specific configuration of the proposed pipeline.}
    \adjustbox{max width=\linewidth}{%
    \begin{tabular}{lccccc}
        \toprule
         & Level 1 & Level 2 & Level 3 & Level 4 & Level 5 \\
        Module & Text-only & Structure & Variable & Image-only & Photography \\
        \midrule
        \textit{Visual Information Extraction} \\
        \cmidrule{1-1}
        Cropper \& Image Sharpening \& Text OCR & \xmark & \xmark & \xmark & \cmark & \xmark \\
        Visual Extractor & \xmark & \xmark & \cmark & \cmark & \xmark \\
        Visual-Grounded Problem Reconstructor & \xmark & \cmark & \cmark & \cmark & \xmark  \\
        \midrule
        \textit{Reasoning} \\
        \cmidrule{1-1}
        Solvers & Multi-agent & Multi-agent & Multi-agent & Multi-agent & Single-agent \\
        Canonicalizer & \cmark & \cmark & \cmark & \cmark & \xmark \\
        Pruner  & \cmark & \cmark & \cmark & \cmark & \xmark \\
        \bottomrule
    \end{tabular}
    }
    \label{tab:module}
\end{table*}

\section{Methodology}
\label{sec:method}

\subsection{Overview}
\label{subsec:method_overview}

We propose an agent-based framework that composes multiple LLM agents across two stages (\emph{Visual Information Extraction} and \emph{Reasoning}) and adapts which modules are active to each level's visual complexity. \cref{fig:overview_level4} shows the full Level~4 pipeline and \cref{tab:module} summarizes the per-level configuration; overviews for the remaining levels appear in \cref{app_sec:overview}.

The \emph{Visual Information Extraction} stage renders a figure into a solver-readable form. For the image-only Level~4, a \emph{Problem Parser} (Cropper, Image Sharpening, Text OCR) first splits the single image into a cleaned diagram and an OCR-derived problem text; Levels~1--3 use the provided image and text directly. A \emph{Visual Extractor} then emits a structured SVG of the diagram (Levels~3--4), and a \emph{Visual-Grounded Problem Reconstructor} rewrites the figure and text into a self-contained problem statement (Levels~2--4). Level~1 is text-only and skips this stage.

The \emph{Reasoning} stage solves the enriched problem with the multi-agent debate pipeline of \cref{subsec:method_reasoning}: three heterogeneous solvers answer independently, a \emph{Canonicalizer} settles the result by value, and, on disagreement, a \emph{Pruner} filters unreliable intermediate solutions before a re-solving round (\cref{subsec:method_reasoning}). Level~5 (photographs) departs from Levels~1--4 and uses a single Gemini solver (\cref{subsec:level5}).

\paragraph{Use of information across levels.} Although Levels~1--4 re-encode the same problems, each evaluation instance is solved using only its own inputs; at inference no information from other levels of the same problem is accessed.

\subsection{Visual Information Extraction}
\label{subsec:method_visual}
\subsubsection{Problem Parser}
At Level~4 each image contains both the problem text and the figure, which degrade extraction when read jointly; the \emph{Problem Parser} therefore separates them into a text region (converted to text) and a diagram region (lightly refined) before downstream use.

\paragraph{Cropper.} Because a bad crop is unrecoverable, the Cropper is a propose--verify loop over \texttt{Gemini-3.5-Flash} (prompts in \cref{app:propose_prompt,app:verify_prompt}). Since the text region is always a contiguous block on the left or top, a single straight boundary suffices: we query the model three times for the boundary orientation, text side, and split fraction (majority vote on the first two, median on the last), then draw the candidate boundary on the image and ask whether the split is clean, moving it as instructed for up to five iterations. Samples that still fail to parse are re-cropped without verification to guarantee an output.

\paragraph{Image Sharpening.} To close the size and quality gap between original Level-3 images and cropped Level-4 diagrams, we refine the crop with deterministic image operations, avoiding an LLM editor, which may distort content. We estimate the background from border pixels, re-crop tightly to the content bounding box, whiten near-white and isolated dark-pixel noise while preserving connected strokes, and apply a mild \texttt{UnsharpMask}. The procedure never resizes or warps the figure, so its geometry is preserved.

\paragraph{Text OCR.} We transcribe the cropped text region into a LaTeX-formatted string with \texttt{Gemini-3.5-Flash} (\cref{app:ocr_prompt}), rejecting outputs that contain meta-commentary rather than the problem statement (``garbage''). For robustness we retry over temperatures $\{0.0, 0.4, 0.7, 0.9\}$, fall back to \texttt{Gemini-2.5-Flash}, and finally accept the transcription with the garbage check disabled.

\subsubsection{Visual Extractor}
The Visual Extractor converts the diagram into a single self-contained SVG that preserves geometry, topology, labels, numerical values, units, arrows, and angles, using \texttt{Gemini-3.5-Flash} (prompt in \cref{app:visual_extractor_prompt}). It reads the level-appropriate inputs: the cropped diagram with OCR text for Level~4, the original image and text otherwise. For reliability it first generates with thinking enabled and a large output budget, falling back to a smaller no-thinking pass if the SVG is incomplete. As post-processing we inline \texttt{<tspan>} sub/superscripts (e.g., \texttt{N\_1}) and cap the SVG at $12{,}000$ characters to bound downstream context.

\paragraph{SVG versus a symbolic parse.} A natural alternative is a plane-geometry diagram parse (PGDP)~\citep{zhang2022pgdp} of primitives and their relations. We adopt SVG for \emph{generality} (it describes arbitrary physics figures (circuits, free-body diagrams, plots), whereas a PGDP is specialized to geometric primitives) and for \emph{coverage and accuracy}: SVG is defined for every question and matches or exceeds a refined PGDP on Level~3 ($0.780$ vs.\ $0.765$; \cref{tab:exp_axis2}). SVG is therefore used at all structure-extracting levels (Levels~3--4).

\subsubsection{Visual-Grounded Problem Reconstructor}
Because reasoning models may not fully exploit raw SVG markup, the Visual-Grounded Problem Reconstructor (\texttt{Gemini-3.5-Flash}; \cref{app:reconstructor_prompt}) rewrites the figure together with the partial problem text into a single, self-contained natural-language problem statement. The prompt enforces strict anti-hallucination: every numerical value, unit, symbol, angle marking, and multiple-choice option is preserved verbatim, no physical relationship or reference frame is inferred, and content that is illegible or ambiguous is flagged rather than guessed.

\subsection{Multi-Agent Debate for Reasoning}
\label{subsec:method_reasoning}

The reasoning stage instantiates the MAD-M$^2$ pipeline of \cref{subsec:prelim_madmm} with three heterogeneous solvers, chosen for their complementary strengths and failure modes: \texttt{GPT-5.5}~\citep{openai2026gpt55} and \texttt{Claude-Opus-4.8}~\citep{anthropic2026opus48} at reasoning-effort \texttt{xhigh}, and \texttt{Gemini-3.1-Pro}~\citep{google2026gemini31} decoded greedily at temperature~$0$. Drawing solvers from different families decorrelates errors (a misreading by one model is unlikely to recur in the others), which is exactly what the memory-masking step exploits. Because these commercial models expose no fixed seed, the pipeline is not bitwise reproducible. \Cref{tab:madmm_pipeline} summarizes the four stages; Level~5 uses a single \texttt{Gemini-3.1-Pro} solver in place of the ensemble.

\paragraph{Solving.} Each solver answers independently from the enriched problem, i.e., the problem text (original or OCR-derived) and, at the SVG levels, the extracted SVG with its figure transcription, and it emits a \texttt{<reasoning>} block followed by an \texttt{<answer>} (option letter(s), or a value with units). This yields one clean candidate per solver, the round-1 memory set $M_1 = \{(a_i^{(1)}, r_i^{(1)})\}_{i=1}^{3}$. Solvers use up to $32{,}768$ output tokens (\texttt{Claude-Opus-4.8}: $32{,}000$); the round-1 prompt is the solving prompt from the SeePhys~Pro baseline of the ICML~2025 SeePhys first-place team~\citep{liang2026scireasoner}, which we use directly and extend only with a multiple-choice answer-format rule (\cref{app:solver_prompt}).

\paragraph{Aggregation and re-solving.} The Canonicalizer (\cref{subsec:canonicalizer}) compares the normalized answers by value and returns a majority directly. Otherwise the Pruner (\cref{subsec:pruner}) masks the unreliable memories to $M_2 = M_1 \otimes \mathit{Mask}$, and the solvers re-solve conditioned only on the surviving memories, for at most two rounds (three solving passes in total). If no majority emerges, a final \texttt{GPT-5.5} step selects the most reliable round-3 candidate, defaulting to the \texttt{Gemini-3.1-Pro} answer if it abstains. Agents never exchange rebuttals directly (each round only re-exposes vetted memories), so unreliable reasoning is suppressed rather than propagated.

\subsubsection{Canonicalizer}
\label{subsec:canonicalizer}
The Canonicalizer denotes the full answer-standardization stage that turns the solvers' raw
outputs into a single, grading-ready final answer; it comprises three steps.
(i)~\emph{Per-agent normalization}: each solver's answer is first cleaned by a lightweight pass (\texttt{gpt-5.4-mini}) that re-extracts the final answer from the solver's reasoning when the emitted answer is malformed or empty; well-formed answers are left unchanged. This yields one clean, comparable answer per solver (the per-solver step in \cref{fig:overview_level4}).
(ii)~\emph{Agreement and false-tie resolution}: the normalized answers are compared by physical value (ignoring formatting, units, notation, ordering, and approximation) and the majority answer is selected. Answers differing only in surface form are recognized as equal by the \texttt{GPT-5.5} value comparison, resolving the apparent tie rather than triggering an unnecessary debate round. (iii)~\emph{Grading-standard rendering}: the agreed answer is rendered once into the
final grading-standard form: preserving the exact value, attaching SI units, using standard
LaTeX, and reducing multiple-choice answers to bare option letters (\texttt{GPT-5.5};
Appendix~\ref{app:canon_prompt}). If no majority emerges, the candidates enter the
prune--mask--re-solve rounds (\cref{tab:madmm_pipeline}).

\subsubsection{Pruner}
\label{subsec:pruner}
Our Pruner realizes the memory-masking step of \cref{subsec:prelim_madmm}. We reuse the
       \texttt{Gemini-3.1-Pro} solver (Solver~2) as a shared Pruner; this is a separate, stateless call to the same model and does not carry over the solver's prior conversation. For efficiency, the Pruner runs at a reduced Gemini thinking budget of $2{,}048$ tokens; empirically, this reduced budget does not change its verdicts relative to full reasoning effort. At the start of each
re-solving round it evaluates every memory once with the pruning prompt
(Appendix~\ref{app:prune_prompt}): grounding its judgment in the figure, it labels
each candidate solution \textsc{yes}, \textsc{no}, or \textsc{not sure} according
to whether the solution is fully correct, paying particular attention to figure
misreadings (wrong topology, swapped labels, or misread values). A subjective mask
is then applied: memories labeled \textsc{yes} are preserved, while \textsc{no}
and, in our strict setting, \textsc{not sure} memories are masked out. Only the
preserved memories are injected into the next round (\cref{tab:madmm_pipeline}); if
the mask removes all of them, the agents re-solve without peer context.

\subsubsection{Level 5: Single-Agent Reasoning}
\label{subsec:level5}
For the photograph-based Level~5 problems we replace the ensemble with a single
\texttt{Gemini-3.1-Pro} solver, using the same model version, greedy decoding
(temperature~$0$), and Round-1 solver prompt (Appendix~\ref{app:solver_prompt}) as the
Gemini solver in the multi-agent setting. Level~5 runs a single solve pass: there is no
debate, pruning, or aggregation, and the solver's answer is taken directly as the final
answer without a separate canonicalization step. We adopt this simplified pipeline because,
empirically, the visual-information-extraction and multi-agent-reasoning modules yielded no
measurable gain on this level.


\section{Experiments}
\label{sec:exp}

We analyze the pipeline along two orthogonal axes that mirror its two stages. The first, \emph{multi-agent reasoning}, improves answer quality once the problem is fully textual and is isolated on the text-only Level~1 (\cref{subsec:axis1}). The second, \emph{closing the modality gap}, converts each figure into a form the solver can read and is studied on Levels~2--4 (\cref{subsec:axis2}). We study each axis on its own before reporting overall performance (\cref{subsec:overall}).

\begin{table*}[t!]
\centering
\caption{Logic of the multi-agent debate pipeline, adopted from MAD-M$^2$.
Each stage targets a different model: the three Solvers (\texttt{GPT-5.5}, \texttt{Gemini-3.1-Pro}, \texttt{Claude-Opus-4.8}), the Canonicalizer, and the Pruner (the Gemini solver acting in a separate
role). $M_1$ is the round-1 memory set, $M_2 = M_1 \otimes \mathit{Mask}$ its pruned version,
and $M_3$ the re-solved set; the superscript on $a_i, r_i$ indexes the memory set. Steps~3--4 run only when step~2 finds no majority and loop back to step~2 for at most two re-solving rounds. If no majority emerges after the final round, a \texttt{GPT-5.5} selection step picks the most reliable of the three final-round candidates, which is then canonicalized.}
\label{tab:madmm_pipeline}
\small
\setlength{\tabcolsep}{5pt}
\renewcommand{\arraystretch}{1.3}
\begin{tabularx}{\linewidth}{@{}l l Y Y@{}}
\toprule
\textbf{Stage} & \textbf{Model} & \textbf{Input} & \textbf{Output} \\
\midrule
1. Solving & Solvers 1--3 & Problem $+$ Solving System Prompt~\citep{liang2026scireasoner} & $M_1 = \{(a_i^{(1)}, r_i^{(1)})\}_{i=1}^{3}$ \\
\midrule
2. Aggregation & Canonicalizer & $a_1, a_2, a_3$ (compared by value) & Majority \cmark{}: canonicalized final answer.\quad Majority \xmark{} (tie): the Canonicalizer resolves it by value: if a shared value emerges it is the final answer; otherwise go to step~3. \\
\midrule
3. Pruning & Pruner & Problem $+\,M_1\,+$ Pruning prompt~\citep{tian2026multiagent} & $M_2 = M_1 \otimes \mathit{Mask}$ \\
\midrule
4. Re-solving & Solvers 1--3 & Problem $+\,M_2\,+$ Debate prompt~\citep{tian2026multiagent} (\cref{app:debate_prompt}) & $M_3 = \{(a_i^{(3)}, r_i^{(3)})\}_{i=1}^{3}$\quad$\circlearrowright$~\emph{repeat from step~2} \\
\bottomrule
\end{tabularx}
\end{table*}

\subsection{Experimental Settings}
\label{subsec:exp_setting}

\paragraph{Benchmark.}
We evaluate on the SeePhys~Pro benchmark (Challenge Track~3)~\citep{xiang2026seephyspro}, which presents the
same physics problems at five levels of increasing visual complexity. Levels~1--4
contain $200$ questions each and Level~5 contains $30$ photograph-based questions,
for $830$ questions in total; this is the public (\emph{testmini}) split, exactly 20\% of each level; the private test ($800$ per level and $120$ for Level~5, $3{,}320$ questions in total) determines the final ranking. We report per-level accuracy (correct$/$level-size)
and the overall accuracy, computed as the micro-average $\mathrm{total\_correct}/830$
(so Level~5 accounts for only ${\sim}3.6\%$ of the overall weight).

\paragraph{Evaluation.}
Submissions must contain both a final answer and supporting reasoning for every
question, and are scored by the organizers' \emph{OpenAI-based semantic judge} against
hidden reference answers\footnote{\url{https://www.codabench.org/competitions/16010/}}:
the judge checks that the submitted answer is equivalent to the reference (respecting
required units) and that the reasoning is relevant and supports it. During development, ground-truth answers were not available locally; every score we
report was obtained from the official organizer leaderboard. Two properties of the
scorer shape our analysis: (i)~it is \emph{format-sensitive} (multiple-choice
answers must be reduced to option letter(s) and numeric answers to a canonical
value/unit form); and (ii)~repeated submissions of
near-identical files vary by roughly one to three questions ($\pm0.005$--$0.015$),
which we treat as noise. We therefore emphasize consistent, sizeable differences
rather than single-point gaps.

\paragraph{Models.}
The reasoning ensemble combines three heterogeneous solvers from different model
families: \texttt{GPT-5.5} and \texttt{Claude-Opus-4.8} at the \texttt{xhigh}
reasoning-effort level and \texttt{Gemini-3.1-Pro} decoded greedily at
temperature~$0$. Visual information extraction (cropping, OCR, SVG generation, and
figure transcription) uses \texttt{Gemini-3.5-Flash}; per-agent answer normalization
uses \texttt{gpt-5.4-mini}; and the canonicalization / value-comparison stage uses
\texttt{GPT-5.5}.

\subsection{Multi-Agent Debate on Level~1 (Axis~1)}
\label{subsec:axis1}
The reasoning stage is isolated on the text-only Level~1, which has no modality gap, so that
any gain is attributable to reasoning alone. Three heterogeneous solvers (\texttt{GPT-5.5},
\texttt{Gemini-3.1-Pro}, \texttt{Claude-Opus-4.8}) answer independently, and a value-based
majority vote forwards only the disagreements to further rounds. Consensus is a reliable and
inexpensive signal: when two solvers agree on a value they are correct ${\sim}83\%$ of the
time, so confident questions terminate early and computation is concentrated on the contested
ones. Drawing the solvers from \emph{different} model families is itself beneficial; replacing
an earlier second \texttt{Gemini} sampler with a different-family solver improved accuracy, as
cross-family disagreement exposes errors that same-family sampling conceals.

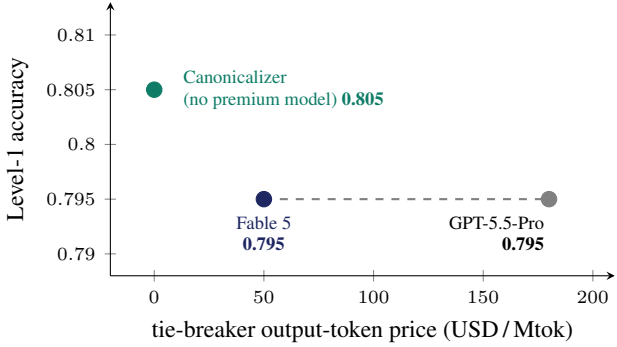
\begin{figure}[!tb]
\centering
\begin{tikzpicture}
\begin{axis}[
  width=\linewidth, height=5.2cm,
  xlabel={tie-breaker output-token price (USD\,/\,Mtok)}, xlabel style={font=\footnotesize},
  ylabel={Level-1 accuracy}, ylabel style={font=\footnotesize},
  xmin=-20, xmax=210, ymin=0.788, ymax=0.813,
  xtick={0,50,100,150,200}, xticklabel style={font=\scriptsize},
  ytick={0.79,0.795,0.80,0.805,0.81},
  yticklabel style={font=\scriptsize,/pgf/number format/fixed,/pgf/number format/precision=3},
  axis lines=left, clip=false,
]
% flat line through the two premium tie-breakers: more price, no gain
\draw[dashed,gray,thick] (axis cs:50,0.795) -- (axis cs:180,0.795);
\addplot[only marks,mark=*,mark size=2.9pt,color=goodg] coordinates {(0,0.805)};
\addplot[only marks,mark=*,mark size=2.9pt,color=navy]  coordinates {(50,0.795)};
\addplot[only marks,mark=*,mark size=2.9pt,color=gray]  coordinates {(180,0.795)};
\node[font=\scriptsize,text=goodg,anchor=west,align=left] at (axis cs:9,0.8050)
  {Canonicalizer\\ (no premium model) \textbf{0.805}};
\node[font=\scriptsize,text=navy,anchor=north,align=center] at (axis cs:50,0.7943)
  {Fable~5\\ \textbf{0.795}};
\node[font=\scriptsize,anchor=north east,align=right] at (axis cs:182,0.7943)
  {GPT-5.5-Pro\\ \textbf{0.795}};
\end{axis}
\end{tikzpicture}
\caption{Paying more per token for the tie-breaker buys no Level-1 accuracy. Both premium
tie-breakers evaluated, \texttt{Fable~5} (\$50/Mtok output) and \texttt{GPT-5.5-Pro}
(\$180/Mtok), score $0.795$, adding zero correct over the debate pipeline, while the
Canonicalizer, which resolves false ties and renders the graded answer format with no
premium model, reaches $0.805$. Output-token prices from \cref{tab:prices}.}
\label{fig:cost}
\end{figure}

\textbf{A premium tie-breaker is unnecessary.} A natural design is to resolve disagreements with a
stronger model, so they were initially routed to \texttt{GPT-5.5-Pro}~\citep{openai2026gpt55pro}.
This tie-breaker did
not improve accuracy over the single solver ($0.720$ in both cases); the gain came instead
from the three-model ensemble itself with value-based selection, which lifts accuracy from
$0.720$ to $0.795$, while additional debate rounds oscillate within grader noise and add no
net correct answers. In the
controlled comparison (the \emph{same} debate pipeline with and without the tie-breaker),
removing it and substituting a lightweight \emph{Canonicalizer} slightly exceeds it:
$\mathbf{0.805}$ without \texttt{GPT-5.5-Pro} versus $0.795$ with it (\cref{tab:exp_axis1},
\cref{fig:cost}).

\textbf{The apparent value of the tie-breaker: false ties and answer format.} The
Canonicalizer performs two inexpensive operations: \emph{value-equivalence voting}, a
majority that compares answers by physical value rather than surface string, and
\emph{grading-standard rendering} into SI units and standard \LaTeX{}, each of which removes
an artifact that had inflated the tie-breaker's apparent contribution. First, naive string
voting mislabels value-equal answers (e.g.\ $4mRv_0/B^2d^2$ and $\tfrac{4mRv_0}{B^2d^2}$) as
three-way ties; approximately $58$ of $63$ ``ties'' were spurious and were escalated
unnecessarily. Resolving ties by value makes genuine ties rare, so the tie-breaker seldom
fires. Second, because the grader is format-sensitive, the tie-breaker's measured gain
reduces to re-rendering an already-correct value into graded form: value-majority on the raw
answers scores $0.745$, standard rendering alone raises it to $0.795$, and rendering the
\emph{same values} in the tie-breaker's own style likewise yields $0.795$. The $+0.050$ is
therefore attributable to formatting rather than reasoning, and the Canonicalizer reproduces
it at no premium cost.

\textbf{The remaining error reflects a capability ceiling.} Substituting an even stronger
tie-breaker (\texttt{Fable~5}~\citep{anthropic2026fable5}) on the hardest contested questions, where \texttt{GPT-5.5-Pro}
had hedged, added \emph{zero} correct answers. This indicates that the ${\sim}0.800$ Level-1
accuracy is a base-capability ceiling rather than a tie-breaking gap
(\cref{app:worked} shows a verbatim instance). The premium models were
never run as standalone full-set solvers, so the claim is limited to their marginal value
\emph{as a tie-breaker}, which is negligible (\cref{sec:limitations}). Removing the
tie-breaker additionally eliminates its ${\sim}\$60$--$70$ per-run cost at no loss in accuracy
(\cref{fig:cost}).

\begin{table}[!tb]
\centering
\caption{Axis~1 on the text-only Level~1: the premium tie-breaker is unnecessary.
\emph{Top block} (controlled comparison): the full debate $+$ Canonicalizer (no Pro) reaches
$\mathbf{0.805}$, above debate $+$ Pro ($0.795$). \emph{Bottom block} (rendering isolated):
holding the answer \emph{values} fixed at the value-majority, the premium model's only
measurable gain is format rendering ($0.745\!\to\!0.795$), which the Canonicalizer reproduces
at no premium cost; the full Canonicalizer's extra point (to $0.805$) comes from the
value-equivalence voting (false-tie resolution) in the top block. All values are public-leaderboard (testmini) accuracy.}
\label{tab:exp_axis1}
\small
\setlength{\tabcolsep}{5pt}
\begin{tabular}{@{}lc@{}}
\toprule
Configuration & Acc. \\
\midrule
\multicolumn{2}{@{}l}{\emph{controlled comparison (debate $\pm$ tie-breaker)}} \\
\quad single solver (\texttt{GPT-5.5}, \texttt{xhigh}) & 0.720 \\
\quad consensus $+$ Pro tie-breaker on disagreements & 0.720 \\
\quad 3-model debate $+$ Pro tie-breaker & 0.795 \\
\quad \textbf{3-model debate $+$ Canonicalizer (no Pro)} & \textbf{0.805} \\
\addlinespace
\multicolumn{2}{@{}l}{\emph{rendering isolated (values fixed at value-majority)}} \\
\quad value-majority, raw answer form & 0.745 \\
\quad $+$ grading-standard rendering only & 0.795 \\
\quad $+$ Pro's rendering only (same values) & 0.795 \\
\bottomrule
\end{tabular}
\end{table}

\subsection{Closing the Modality Gap (Axis~2)}
\label{subsec:axis2}
With the reasoning stack of \cref{subsec:method_reasoning} held fixed, only the
\emph{visual front-end} (the component that converts a figure into a solver-readable
form) is varied. The central empirical finding is that the value of a figure aid scales
with how much of the problem is locked inside the image, i.e.\ with the \emph{modality gap}.
\Cref{tab:exp_axis2} reports the evidence; all configurations share the same
debate-and-canonicalization solver and differ only in the figure representation.
Worked examples of these effects, with verbatim model outputs, are collected in
\cref{app:worked}.

\begin{table*}[t!]
\centering
\caption{Axis~2 (modality-gap) ablation: only the visual front-end is varied; the reasoning
stack is held fixed. All values are public-leaderboard (testmini) accuracy. Where repeated,
near-identical submissions of the same front-end differed within grader noise, we report the
best observed score: the noisy grader can erroneously reject a correct answer but does not
accept an incorrect one, so the maximum best reflects the front-end's accuracy. The gain
tracks the modality gap (L4 $\gg$ L2 $>$ L3).}
\label{tab:exp_axis2}
\small
\begin{tabular}{@{}l l c@{}}
\toprule
Level & Visual front-end (figure representation) & Acc. \\
\midrule
\multirow{3}{*}{\shortstack[l]{L2\\ \emph{structure}}}
 & image $+$ text, no aid & 0.790 \\
 & $+$ SVG & 0.800 \\
 & $+$ transcription (no SVG) & \textbf{0.830} \\
\midrule
\multirow{5}{*}{\shortstack[l]{L3\\ \emph{text-redundant}}}
 & image $+$ text, no aid (bare) & 0.755 \\
 & figure$\to$text \,/\, transcription-only (no image) & 0.755 \\
 & $+$ transcription $+$ image & 0.780 \\
 & $+$ transcription $+$ SVG $+$ image (transsvg) & \textbf{0.780} \\
 & refined PGDP parse $+$ image & 0.765 \\
\midrule
\multirow{6}{*}{\shortstack[l]{L4\\ \emph{image-only}}}
 & figure dropped (image-handling bug) & 0.000 \\
 & original image only (baseline) & 0.540 \\
 & figure$\to$text, no image & 0.710 \\
 & figure-text $+$ image & 0.735 \\
 & transcription $+$ SVG $+$ image (transsvg) & 0.765 \\
 & crop $+$ OCR $+$ SVG $+$ transcription & \textbf{0.775} \\
\bottomrule
\end{tabular}
\end{table*}

\textbf{Level~2 (structure): a prose transcription is the most effective aid.} At Level~2
the pipeline uses the figure transcription alone, without SVG. The transcription raises
accuracy from $0.790$ (image$+$text) to $\mathbf{0.830}$, the best Level-2 result, whereas an
SVG-only front-end reaches $0.800$; the improvement is therefore attributable to the
transcription rather than the SVG. Level~2 provides the cleanest single-aid evidence, since
by construction no other figure representation is active at this level.

\textbf{Level~4 (image-only): extraction is the bottleneck.} Level~4 carries \emph{no}
problem text (the entire question resides in the image), so the figure must be recovered
before reasoning can begin. When an image-handling bug silently dropped the figure, Level~4
accuracy collapsed to $0$ (near-total refusal); restoring the image recovered $0.540$.
Notably, the \emph{raw} image alone (``bare'') underperforms a \emph{textual} rendering of
the same figure: on the subset where Levels~1 and~2 agree, the textual rendering matches
$112/149$ questions against $106/149$ for the bare image ($-6$). The figure is therefore
converted into a textual or structured form rather than read directly, and extraction
fidelity is then increased progressively. Each step raises accuracy roughly monotonically:
figure$\to$text $0.710 \rightarrow$ text$+$image $0.735 \rightarrow$
transcription$+$SVG$+$image $0.765 \rightarrow$ crop$+$OCR$+$SVG$+$transcription
$0.775$, a net $+0.235$ over the raw-image baseline. The final Level-4 front-end
crops the page into separate text and diagram regions (OCR for the former, SVG and a
natural-language transcription for the latter) so that embedded text and figure no longer
corrupt each other.

\textbf{Level~3 (text-redundant): figure aids fall within grader noise.} At Level~3 the
figure largely duplicates information already present in the text, and the results form a
flat plateau: three independent, generalizable front-ends (no aid, figure$\to$text, and
transcription-only) all score \emph{exactly} $0.755$. Injecting the image at the
answer-resolution stage changes only $23/200$ predictions and is neutral-to-harmful,
confirming that the Level-3 image is effectively redundant. A cumulative comparison (image
retained) gives no-aid $0.755 \rightarrow {+}$transcription $0.780 \rightarrow {+}$SVG
($=$transsvg) $0.780$, so transcription is a marginal lever and the added SVG sits within
grader noise. A refined symbolic diagram parse (refined PGDP; see \cref{subsec:method_visual})
scores $0.765$, below the SVG-based transsvg.

\textbf{The figure-aid gain scales with the modality gap.} Across levels the gain tracks the
gap directly: Level~4 (image-only) $+0.235 \gg$ Level~2 (structure; transcription
$0.790\!\rightarrow\!0.830$, $+0.040$) $>$ Level~3 (text-redundant; $\approx\!0$). Level~3's
flatness is thus the low-gap end of a single trend rather than a failure of the method.

\textbf{Caveats.} All scores come from the organizer leaderboard (no local ground truth)
and carry $\pm1$--$3$ question ($\pm0.005$--$0.015$) grader noise. The claims therefore rest
on the large cross-level pattern and the Level-4 ladder rather than on fine
per-representation rankings, and a fully controlled aid$\times$aid factorial was not run.

\subsection{Overall Performance}
\label{subsec:overall}
Combining the per-level configurations, \cref{tab:exp_main} reports the full pipeline
against a single-pass \texttt{GPT-5.5} baseline. The framework improves every level, raising
overall accuracy from $0.643$ to $\mathbf{0.802}$, with the largest gains on the image-only
Level~4 ($0.540\!\rightarrow\!0.775$) and the photograph Level~5 ($0.670\!\rightarrow\!0.933$).
Level~5 uses a single \texttt{Gemini-3.1-Pro} solver: the three-agent ensemble degrades to
${\sim}0.500$ because, for options drawn inside the figure, \texttt{GPT-5.5} and \texttt{Claude}
report the computed \emph{value} rather than the option \emph{letter} and outvote Gemini's
correct letters, so a single solver avoids the format clash and reaches $0.933$ (a verbatim instance
appears in \cref{app:worked}). The submitted system won 1st place on both the public and the private leaderboard of
Track~3; on the private test that determines the final ranking it scores $0.743$ overall
($0.745/0.765/0.730/0.713/0.875$ on L1--L5).

\begin{table}[b]
\centering
\caption{Per-level and overall accuracy on SeePhys~Pro (organizer leaderboard). The top
two rows are public (testmini) scores; the bottom row is the hidden private test, which
determines the final ranking (1st place).}
\label{tab:exp_main}
\small
\setlength{\tabcolsep}{4pt}
\adjustbox{max width=\linewidth}{%
\begin{tabular}{lcccccc}
\toprule
System & L1 & L2 & L3 & L4 & L5 & Overall \\
\midrule
GPT-5.5 single-pass & 0.720 & 0.660 & 0.650 & 0.540 & 0.670 & 0.643 \\
Full pipeline (ours) & \textbf{0.805} & \textbf{0.830} & \textbf{0.780} & \textbf{0.775} & \textbf{0.933} & \textbf{0.802} \\
\midrule
Full pipeline, private test & 0.745 & 0.765 & 0.730 & 0.713 & 0.875 & 0.743 \\
\bottomrule
\end{tabular}}
\end{table}

\subsection{Cost and Efficiency}
\label{subsec:cost}
Accuracy was pursued under an explicit efficiency budget rather than through brute-force
scaling; \cref{tab:exp_cost} summarizes the main levers, each of which holds or improves
accuracy while reducing cost or latency. \textbf{Dropping the premium tie-breaker} removes
its ${\sim}\$60$--$70$ per-run cost, since \texttt{GPT-5.5-Pro} is priced at \$30/\$180 per Mtok
(input/output), far above the standard solvers (\cref{tab:prices}); the Canonicalizer
matches and slightly exceeds it. \textbf{Reducing the pruner's reasoning effort} to a
$2{,}048$-token thinking budget reduces its output tokens by $5$--$9\times$ and its per-call
latency from ${\sim}18$ to ${\sim}1$ minute, with \emph{identical} prune verdicts ($9/9$ on
the probe set). \textbf{Early-stopping on consensus} concentrates computation on the
disagreements, saving roughly half of the debate cost at no measurable accuracy loss.
\textbf{Capping the debate at two rounds} avoids the linear per-round cost of additional
rounds, which yield zero or negative gain. Finally, output (reasoning) tokens dominate cost
(${\sim}91\%$), so prompt caching is not a useful lever in this setting. The strongest single
design choice (removing the premium tie-breaker) therefore reduces cost and improves
accuracy simultaneously.

\begin{table}[!tb]
\centering
\caption{API token prices used for cost accounting (USD per million tokens).
Output (reasoning) tokens dominate spend (${\sim}91\%$), so the output rate sets the
per-run total. The two premium models were used only as tie-breakers on contested
questions; \texttt{GPT-5.5-Pro}'s \$180 output rate is what made routing ties to it the
dominant cost.}
\label{tab:prices}
\small
\setlength{\tabcolsep}{7pt}
\begin{tabular}{@{}lcc@{}}
\toprule
Model & Input & Output \\
 & (\$/Mtok) & (\$/Mtok) \\
\midrule
\multicolumn{3}{@{}l}{\emph{solvers}} \\
\quad GPT-5.5            & 5  & 30 \\
\quad Gemini-3.1-Pro     & 2  & 12 \\
\quad Claude-Opus-4.8    & 5  & 25 \\
\addlinespace
\multicolumn{3}{@{}l}{\emph{tie-breakers (contested questions only)}} \\
\quad Fable~5            & 10 & 50 \\
\quad GPT-5.5-Pro        & 30 & \textbf{180} \\
\bottomrule
\end{tabular}
\end{table}

\begin{table}[!tb]
\centering
\caption{Efficiency levers on the reasoning stage. Each cuts cost or latency while holding or
improving leaderboard accuracy; the last column reports the resulting accuracy change ($\Delta$).}
\label{tab:exp_cost}
\small
\setlength{\tabcolsep}{4pt}
\adjustbox{max width=\linewidth}{%
\begin{tabular}{@{}lll@{}}
\toprule
Design decision & Cost / latency effect & $\Delta$ Acc. \\
\midrule
Drop the \texttt{GPT-5.5-Pro} tie-breaker & saves \$60--70 / run & $+0.010$ \\
Pruner effort $\to$ $2{,}048$ tokens & $5$--$9\times$ fewer tokens; $18\!\to\!1$\,min & $0.000$~({\scriptsize $9/9$ verdicts}) \\
Early-stop on consensus & ${\sim}$half the debate compute & ${\approx}0.000$ \\
Cap debate at two rounds & avoids linear per-round cost & ${\approx}0.000$ \\
\bottomrule
\end{tabular}}
\end{table}

% \section{Related Work}
% \label{sec:related_work}

\section{Limitations}
\label{sec:limitations}
Our study has several limitations. (i)~\emph{Leaderboard-only evaluation.} We had no local
ground truth; all scores come from the organizer leaderboard and carry ${\sim}1$--$3$
question grader noise, so we rely on consistent trends rather than single-point gaps.
(ii)~\emph{Tie-breaker, not standalone, comparison.} The premium models
(\texttt{GPT-5.5-Pro}, \texttt{Fable~5}) were only ever used as tie-breakers / re-solvers on
contested questions, never as standalone full-set solvers. We therefore claim that their
marginal value \emph{as a tie-breaker} on top of our pipeline is negligible, \emph{not}
that our pipeline outperforms them as standalone systems; a cost-matched standalone
comparison is left to future work. (iii)~\emph{Incomplete factorial.} We did not run a fully
controlled aid$\times$aid factorial for the visual front-ends, and several per-step modality
deltas sit at the noise floor. (iv)~\emph{Reproducibility.} The commercial models expose no
fixed seed and decode by reasoning effort, so outputs are not bitwise reproducible.

\section{Conclusion}
\label{sec:conclusion}

We presented a two-stage pipeline, visual information extraction followed by
multi-agent reasoning, for visual physics problem solving and used the same-problem-across-levels design of SeePhys Pro to study it
along two orthogonal axes. On the reasoning axis, isolated on the text-only Level~1,
the return from orchestrating heterogeneous solvers came from reliable and inexpensive
answer selection (consensus-based early termination and value-aware canonicalization)
rather than from additional debate; a premium model used as the adjudicator added cost
but no correct answers once value-equal ties were resolved, and accuracy was bounded by
base-model capability. On the modality axis, holding the reasoning stack fixed and
varying only the visual front-end, the value of a figure aid scaled with the fraction
of the problem locked in the image: large on the image-only Level~4, where visual
extraction rather than reasoning is the bottleneck, and negligible where the figure
merely duplicates the text.

Combining consensus-routed debate with modality-adaptive extraction, and extending
caption-assisted reasoning to a structured SVG representation, the full pipeline
improved over a single-pass baseline from $0.643$ to $0.802$ overall on the public split
and finished 1st on both the public and the private leaderboard of Challenge Track~3. The broader lesson
is to spend orchestration where it pays: cheap, value-aware selection on the reasoning
side, and faithful figure extraction on the perception side. As all scores come from
the organizer leaderboard, we report consistent cross-level trends rather than
single-point gaps; a cost-matched comparison against premium models as standalone
solvers, and a fully controlled aid-by-aid factorial, remain natural directions for
future work.

\section*{Acknowledgments}
This work was supported in part by the Institute of Information \& Communications Technology Planning \& Evaluation (IITP)-Information Technology Research Center (ITRC) grant funded by the Korea government (MSIT) (IITP-2026-RS-2024-00437268) (50\%), and in part by the IITP grant funded by the Korea government (MSIT) (No. RS-2022-II220077, AI Technology Development for Commonsense Extraction, Reasoning, and Inference from Heterogeneous Data) (50\%).

\bibliography{example_paper}
\bibliographystyle{icml2026}

%%%%%%%%%%%%%%%%%%%%%%%%%%%%%%%%%%%%%%%%%%%%%%%%%%%%%%%%%%%%%%%%%%%%%%%%%%%%%%%
%%%%%%%%%%%%%%%%%%%%%%%%%%%%%%%%%%%%%%%%%%%%%%%%%%%%%%%%%%%%%%%%%%%%%%%%%%%%%%%
% APPENDIX
%%%%%%%%%%%%%%%%%%%%%%%%%%%%%%%%%%%%%%%%%%%%%%%%%%%%%%%%%%%%%%%%%%%%%%%%%%%%%%%
%%%%%%%%%%%%%%%%%%%%%%%%%%%%%%%%%%%%%%%%%%%%%%%%%%%%%%%%%%%%%%%%%%%%%%%%%%%%%%%
\newpage
\appendix
\onecolumn
\section{Pipeline Overview by Level}
\label{app_sec:overview}

\begin{figure}[ht]
    \centering
    \includegraphics[width=0.94\linewidth]{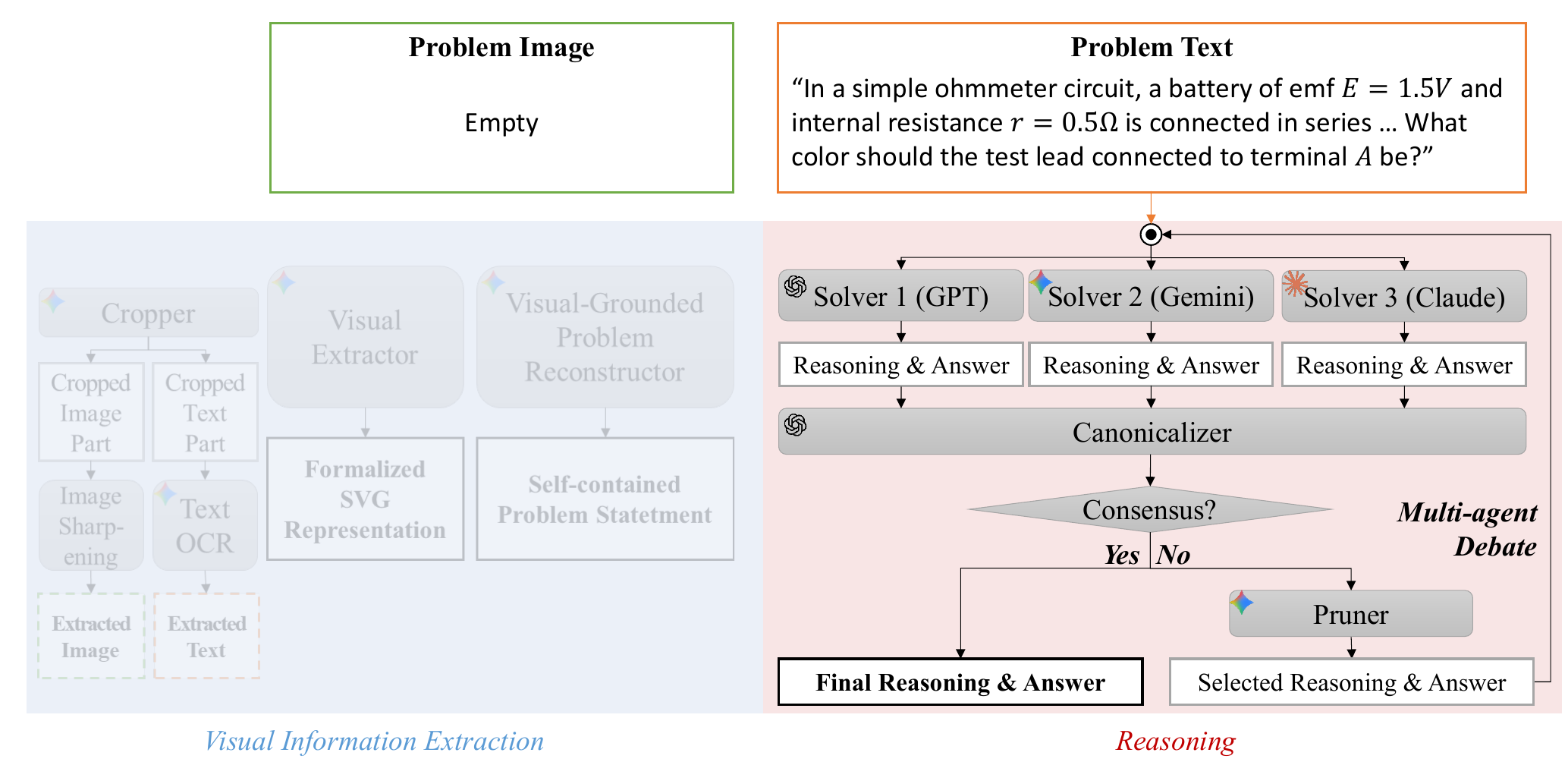}
    \caption{Proposed pipeline for text-only reasoning on Level 1 problems in the SeePhys Pro dataset.}
    \label{fig:overview_level1}
\end{figure}

\begin{figure}[ht]
    \centering
    \includegraphics[width=0.94\linewidth]{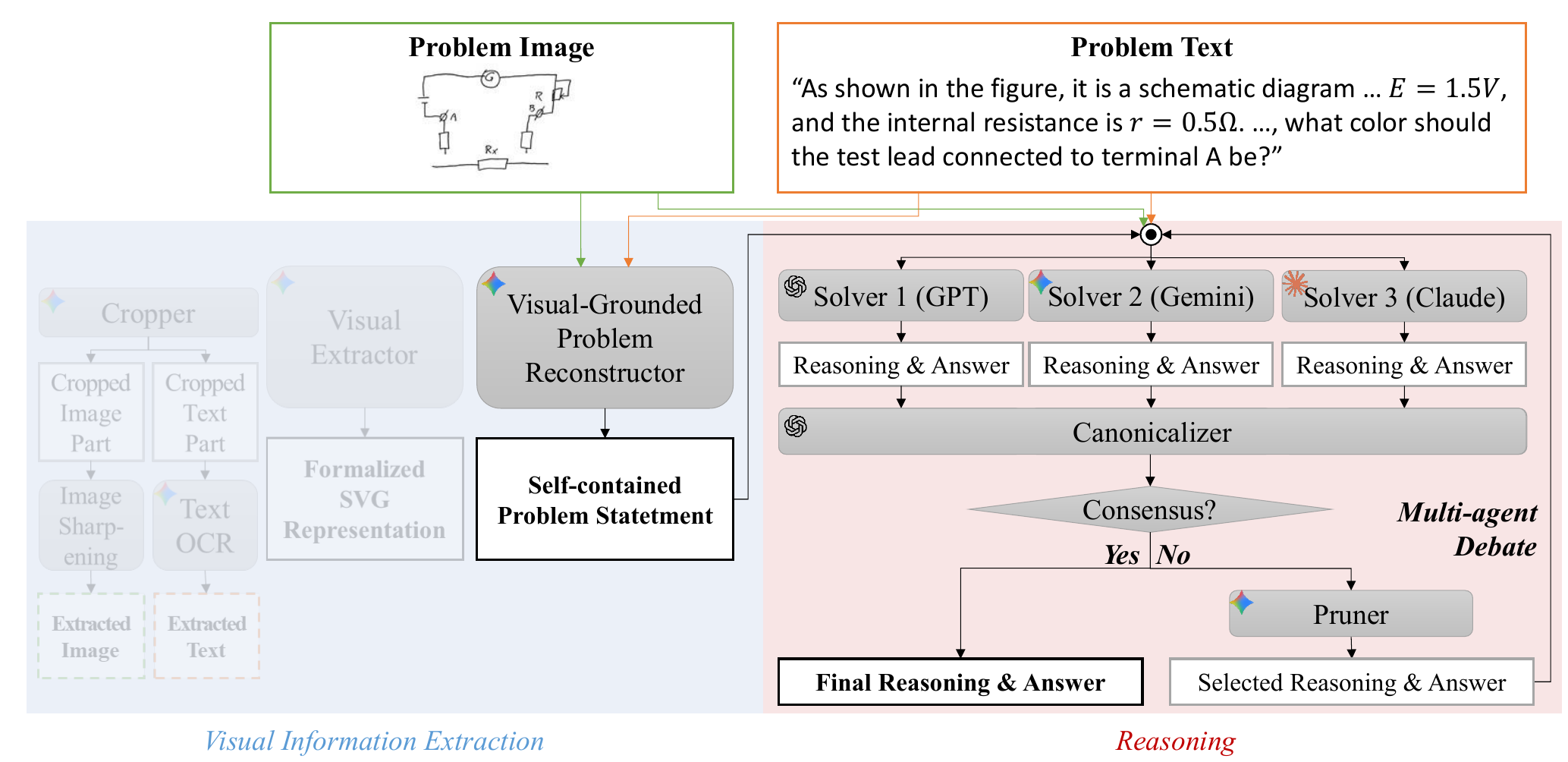}
    \caption{Proposed pipeline for structure-grounded visual reasoning on Level 2 problems in the SeePhys Pro dataset.}
    \label{fig:overview_level2}
\end{figure}

\begin{figure*}[ht]
    \centering
    \includegraphics[width=\linewidth]{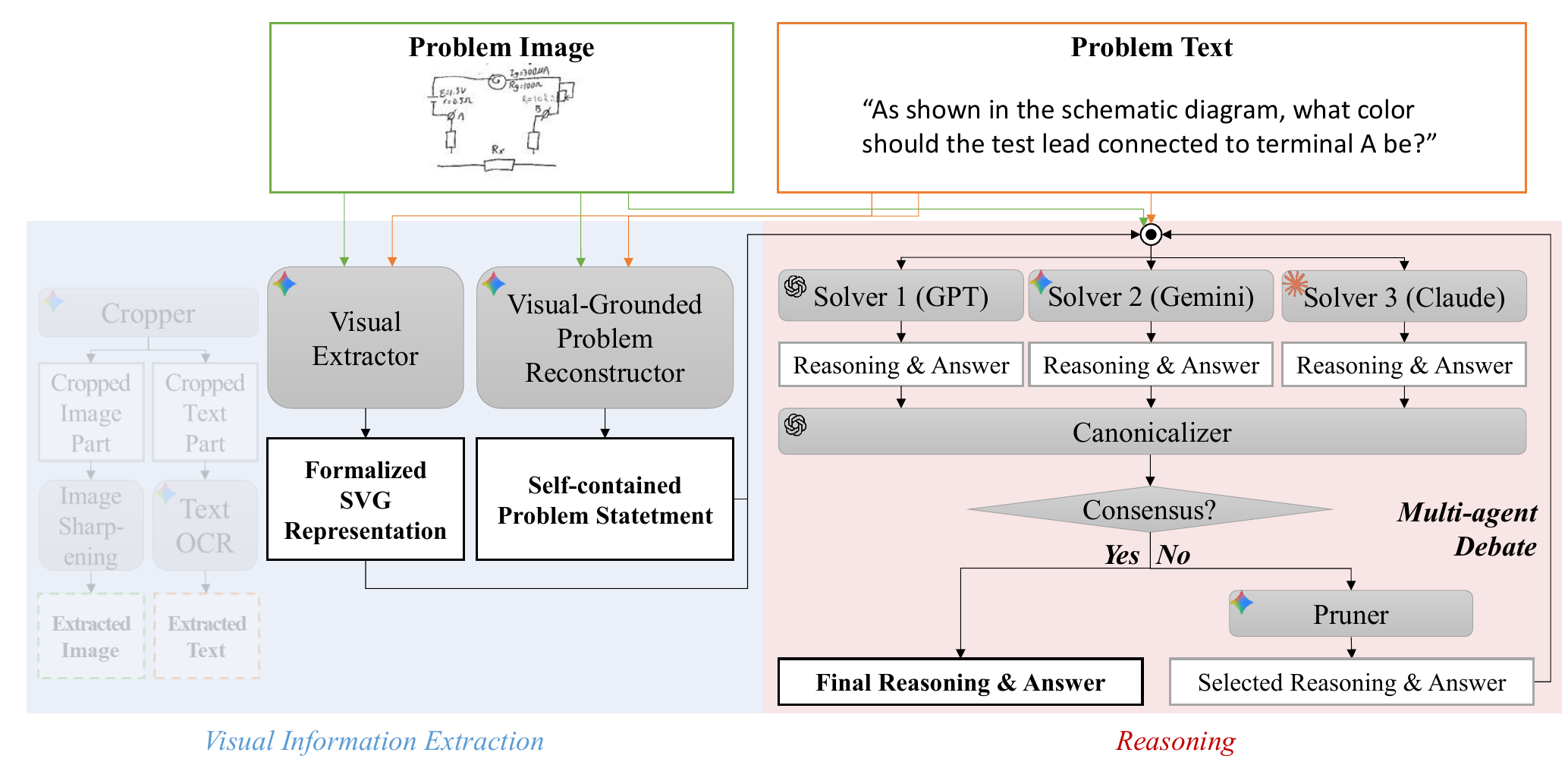}
    \caption{Proposed pipeline for variable-grounded visual reasoning on Level 3 problems in the SeePhys Pro dataset.}
    \label{fig:overview_level3}
\end{figure*}

\begin{figure*}[ht]
    \centering
    \includegraphics[width=\linewidth]{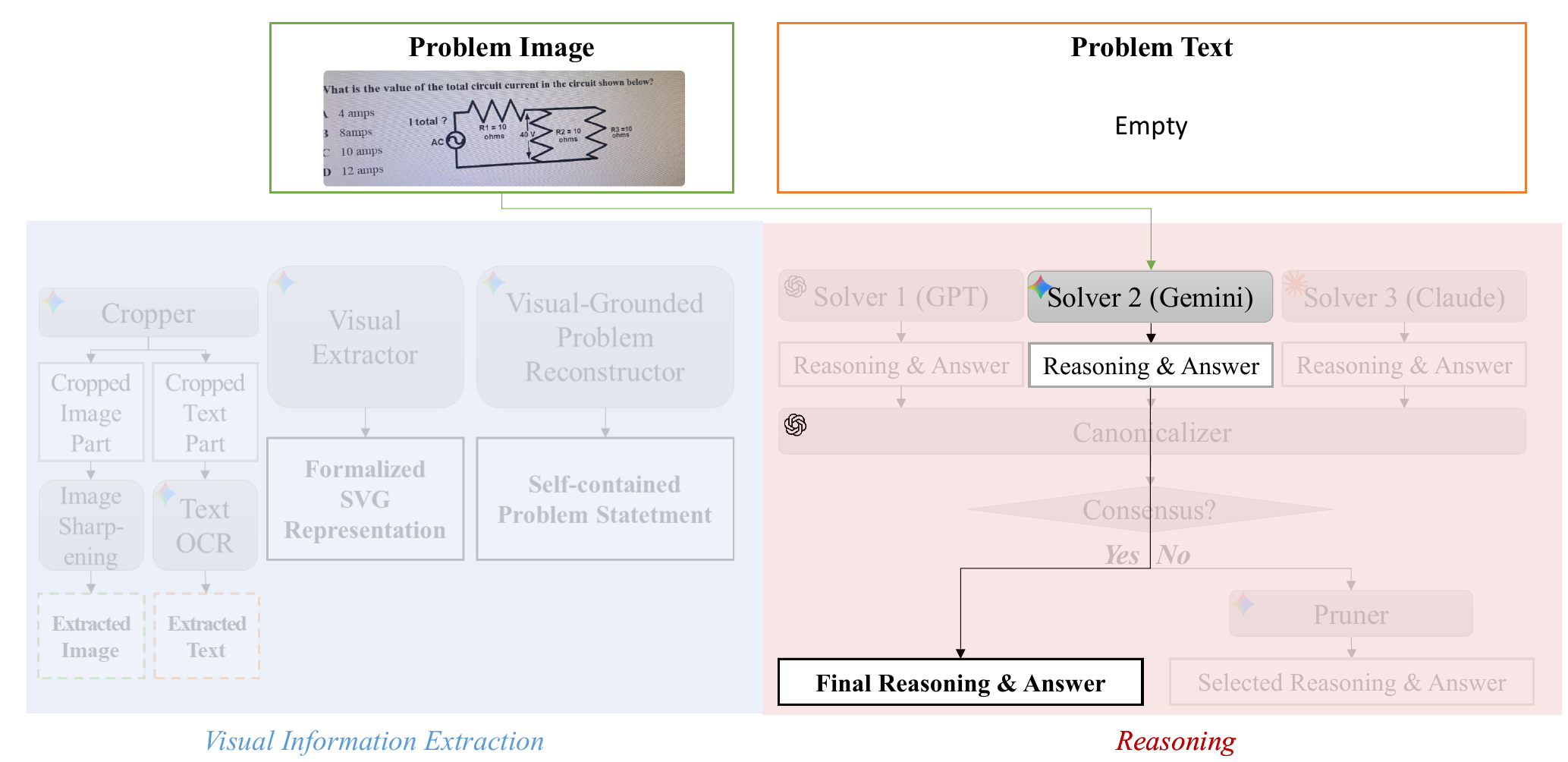}
    \caption{Proposed pipeline for photo-based visual reasoning on Level 5 problems in the SeePhys Pro dataset.}
    \label{fig:overview_level5}
\end{figure*}

This section provides pipeline overviews for the remaining levels of the SeePhys Pro dataset.
\cref{fig:overview_level1} shows the Level 1 pipeline, which performs text-only reasoning. \cref{fig:overview_level2} presents the Level 2 pipeline, where the Visual-Grounded Problem Reconstructor incorporates structural visual information into a self-contained problem statement. \cref{fig:overview_level3} shows the pipeline for Level 3, which additionally employs the Visual Extractor to capture variable-level information from the diagram and generate a structured SVG representation. Finally, \cref{fig:overview_level5} depicts the Level 5 pipeline, which adopts a simplified Gemini-based single-agent reasoning framework for photograph-based questions.

% \appendix

\clearpage        % 현재 페이지 정리
% \onecolumn        % 한 단으로 전환 (이후 전체가 1단)

\section{Worked Examples}
\label{app:worked}
This appendix collects the concrete cases behind the findings of \cref{subsec:axis1,subsec:axis2,subsec:level5}. Model outputs are quoted verbatim from our runs; problem inputs are shown as given to the solvers (abridged where marked with ``$\ldots$'').

\newtcolorbox{inputbox}[1]{enhanced jigsaw, breakable, width=0.98\linewidth,
  colback=blue!3, colframe=gray!60, title={#1}, fonttitle=\bfseries, pad at break*=1mm}
\newtcolorbox{outputbox}[1]{enhanced jigsaw, breakable, width=0.98\linewidth,
  colback=gray!5, colframe=gray!60, title={#1}, fonttitle=\bfseries, pad at break*=1mm}

\subsection{A Reasoning Failure on Text-Only Level 1}
\label{app:worked_l1}
On \texttt{level1\_testmini\_000087} (text-only: a block leaves a rough incline, $\theta = 37^\circ$, $\mu = 0.3$, and passes through a slot in a platform at height $2R$ rotating at angular velocity $\omega$), \texttt{GPT-5.5} at \texttt{xhigh}, solving alone, derives the correct answer and then argues itself out of it:

\begin{outputbox}{GPT-5.5 (xhigh), solving alone --- verbatim excerpts}
\small
``First trip from $A$ to $C$: \(v_1^2 = \tfrac{14}{5}gR\), \quad \(u_1^2 = v_1^2 - 2gR = \tfrac{4}{5}gR\). \\
The time between upward and downward crossings of the platform is \(\Delta t_1 = \tfrac{2u_1}{g} = \tfrac{4}{\sqrt{5}}\sqrt{R/g}\). \\
If this corresponded to one full revolution, then \(\omega = \tfrac{2\pi}{\Delta t_1} = \tfrac{\pi\sqrt{5}}{2}\sqrt{g/R}\).'' \hfill \textcolor{teal2}{\textbf{$\leftarrow$ the correct answer}}

\medskip
``However, we must check the later stated condition. $\ldots$ the slider cannot reach the platform a second time at all. The stated conditions are mutually inconsistent, so \textbf{no angular velocity can satisfy the full problem} as written.''

\medskip
Submitted answer: \textbf{``No physical solution''} \hfill \textcolor{red}{\textbf{$\leftarrow$ rejects its own correct value}}
\end{outputbox}

The answer existed in the candidate pool; the failure was in settling on it, which is exactly what the value-aware selection of \cref{subsec:axis1} repairs.

\subsection{A Perception Failure on Level 3 and Its Recovery}
\label{app:worked_l3}

\begin{inputbox}{Input (Level 3, \texttt{level3\_testmini\_000013}) --- problem text and figure}
\small
``As shown in the figure, the coil has resistance $r$, and each turn has area $S$. The coil supplies power to the external resistor through slip rings and brushes. Regarding this power generation process, the following statements are correct: (\ \ ) \\
$\ldots$ \quad B. The peak value of the induced electromotive force is $40\,\mathrm{V}$ \quad $\ldots$ \quad D. The average electric power on resistor $R$ is $9\,\mathrm{W}$''
\begin{center}
  \includegraphics[width=0.44\linewidth]{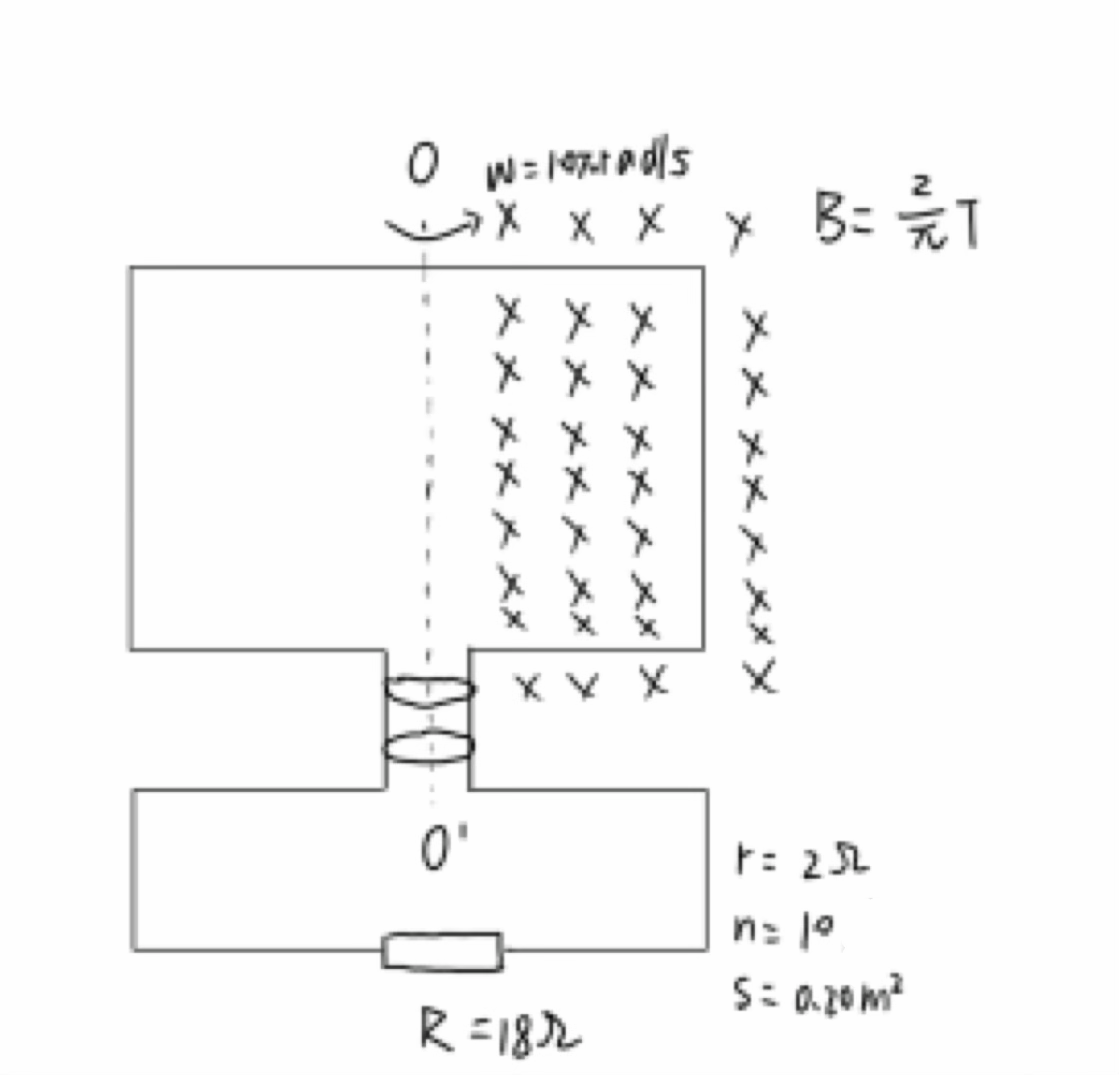}
\end{center}
\emph{The decisive fact exists only in the figure: the field region ($\times$ marks) covers only the half-plane to the right of the rotation axis $OO'$.}
\end{inputbox}

\begin{outputbox}{Claude-Opus-4.8 (xhigh), raw image --- verbatim excerpt}
\small
``Peak EMF: \(\varepsilon_0 = nBS\omega = 10 \times (2/\pi) \times 0.20 \times 10\pi = 40\,\mathrm{V}\). \\
Average power on $R$: \(I_\mathrm{rms} = \sqrt{2}\,\mathrm{A}\); \(P_R = (\sqrt{2})^2 \cdot 18 = 36\,\mathrm{W} \; (\neq 9\,\mathrm{W})\).''
\hfill \textcolor{red}{\textbf{misses the half-region field $\rightarrow$ wrong}}
\end{outputbox}

With the half-region constraint recovered, the effective area halves: $\varepsilon_0 = 20$\,V and $P_R = 9$\,W (choice D). Given the prose transcription, all three solvers stated the constraint in round~1 and agreed on D immediately.

\subsection{Extraction Flips the Vote on Image-Only Level 4}
\label{app:worked_l4}

\begin{inputbox}{Input (Level 4, \texttt{level4\_testmini\_000002}) --- a single image; the problem-text field is empty}
\begin{center}
  \includegraphics[width=0.8\linewidth]{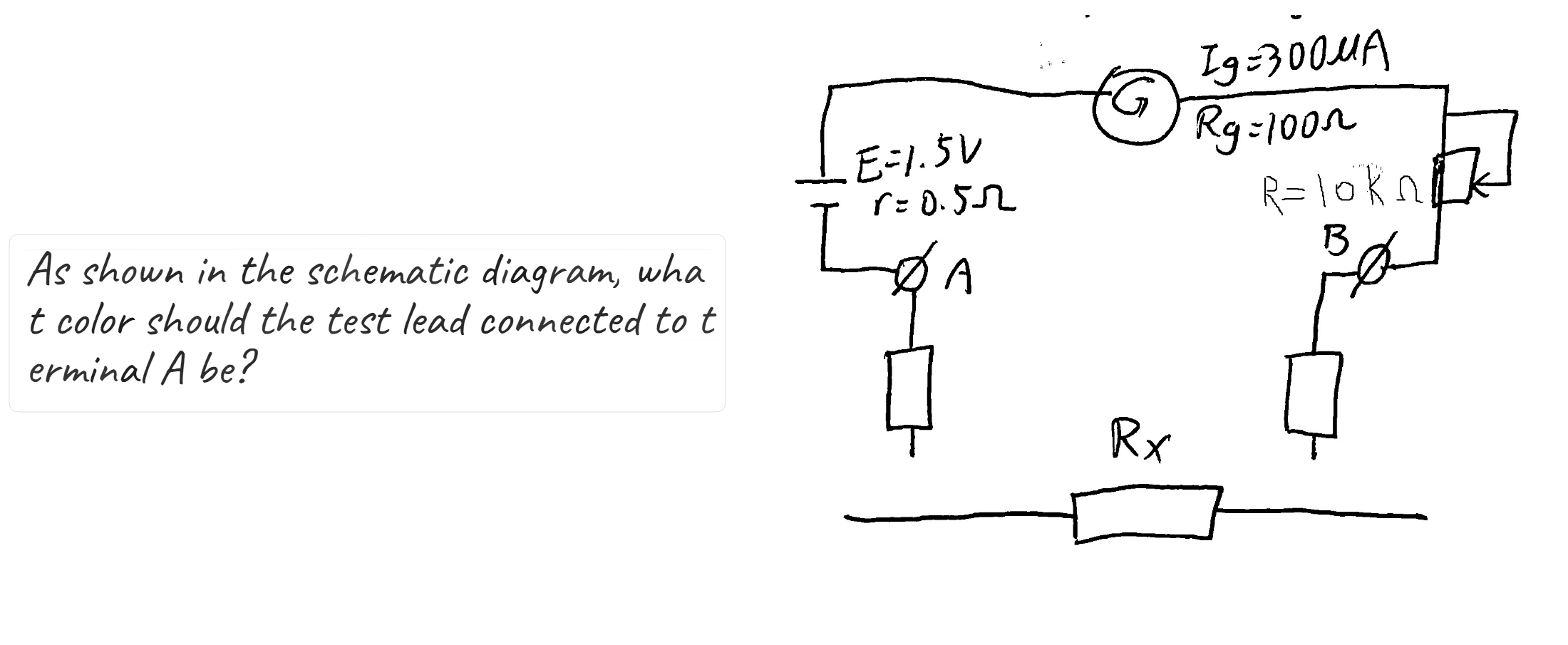}
\end{center}
\small
\emph{The answer hinges on the internal battery's polarity at terminal $A$, a fact that exists only in the figure. The extraction stage recovers it in the transcription --- ``The \textbf{negative terminal} of the battery is connected to \textbf{terminal $A$}$\,\ldots$'' --- and the structured SVG encodes the same fact in its markup.}
\end{inputbox}

\begin{outputbox}{Claude-Opus-4.8 (xhigh) --- verbatim excerpts, before and after extraction}
\small
\textbf{Raw image only:} ``In resistance mode, the internal battery's POSITIVE terminal is connected to the COM jack $\rightarrow$ black lead$\ldots$ Since the lead connected to the battery's negative side in ohm mode is the red lead, \textbf{terminal $A$'s test lead must be red}.''
\hfill \textcolor{red}{\textbf{$\rightarrow$ Red; majority among the three solvers: Red}}

\medskip
\textbf{With extraction:} ``Terminal $A$ is connected to the \textbf{negative} terminal of the battery$\ldots$ By multimeter color convention, the negative/common probe is \textbf{black}$\ldots$ its lead should be black.''
\hfill \textcolor{teal2}{\textbf{$\rightarrow$ Black; majority among the three solvers: Black}}
\end{outputbox}

Same model, same question --- only the input changed; the majority answer among the three solvers flips from \emph{Red} to the correct \emph{Black}.

\subsection{Level 5: the Letter-versus-Value Clash}
\label{app:worked_l5}
On \texttt{solve150\_testmini\_000020} (a photograph: a $6$\,V battery drives a $1.25\,\Omega$ load; the options are printed inside the photo and the prompt requires the option letter):

\begin{outputbox}{GPT-5.5 vs.\ Gemini-3.1-Pro --- verbatim}
\small
\textbf{GPT-5.5:} ``\(P = V^2/R = 6^2/1.25 = 28.8\). So the power supplied is $28.8$\,W, \textbf{corresponding to option D}.'' \quad Submitted answer: \textbf{``28.8''}
\hfill \textcolor{red}{\textbf{names the letter, submits the value}}

\medskip
\textbf{Gemini-3.1-Pro:} ``$\ldots$ So the correct option is D.'' \quad Submitted answer: \textbf{``D''}
\hfill \textcolor{teal2}{\textbf{correct letter}}
\end{outputbox}

Under value voting the two value-format answers outvote the correct letter, which is why Level~5 uses a single Gemini solver (\cref{subsec:level5}).

\section{Full Prompts}
\subsection{Cropper Prompt}
\label{app:propose_prompt}
\begin{tcolorbox}[
    enhanced jigsaw,
    breakable,
    width=0.98\linewidth,
    colback=gray!5,
    colframe=gray!60,
    title=Prompt for Boundary Proposal,
    fonttitle=\bfseries,
    pad at break*=1mm,
]
\scriptsize
\begin{Verbatim}[breaklines=true,breakanywhere=true]
This image is a physics problem. It contains a printed
problem statement with multiple-choice options (the "text" region) and a
figure/diagram/plot (the "diagram" region). They sit side by side or stacked.

Decide the single straight line that best separates the two regions, and tell me
which side holds the problem text. Return ONLY this JSON (no markdown):

{
  "orientation": "vertical" | "horizontal",
  "text_side": "left" | "right" | "top" | "bottom",
  "split_fraction": 0.0
}

- "vertical": the separating line is vertical (text and diagram are left/right).
- "horizontal": the line is horizontal (text and diagram are top/bottom).
- "split_fraction": where the line sits along the relevant axis, 0..1
  (for vertical = x fraction from the left; for horizontal = y fraction from top).
- A rough estimate is fine; it will be checked and refined afterwards.
- Short labels inside the figure (A, B, R1, M, N) belong to the DIAGRAM.
\end{Verbatim}
\end{tcolorbox}

\subsection{Cropper Verification Prompt}
\label{app:verify_prompt}
\begin{tcolorbox}[
    enhanced jigsaw,
    breakable,
    width=0.98\linewidth,
    colback=gray!5,
    colframe=gray!60,
    title=Prompt for Boundary Verification,
    fonttitle=\bfseries,
    pad at break*=1mm,
]
\scriptsize
\begin{Verbatim}[breaklines=true,breakanywhere=true]
You are a strict QA checker for an image-cropping pipeline.

The attached image is a physics problem. A RED line has been drawn to split it
into two parts: one part should contain ONLY the printed problem statement and
its multiple-choice options (the "text" region), and the other part should
contain ONLY the figure/diagram/plot (the "diagram" region).

Short axis labels that belong to the figure (like A, B, R1, M, N) count as part
of the DIAGRAM, not as problem text.

Judge whether the red line splits these two regions CLEANLY. Return ONLY this
JSON (no markdown, no commentary):

{
  "clean": true | false,
  "text_region": "left" | "right" | "top" | "bottom",
  "issue": "none" | "text_truncated" | "diagram_truncated" | "diagram_leaks_into_text" | "text_leaks_into_diagram",
  "move": "none" | "toward_text" | "toward_diagram",
  "move_fraction": 0.0
}

Definitions:
- "clean": true only if the problem text/options are entirely on one side AND
  the figure is entirely on the other side, with nothing important cut by the line.
- issue "text_truncated": some problem text/options got cut and ended up on the
  diagram side  -> the line must move TOWARD the diagram to give text more room.
- issue "diagram_leaks_into_text": part of the figure is on the text side
  -> the line must move TOWARD the text.
- "move": which direction the line should shift to fix it. "none" if clean.
- "move_fraction": how far to move, as a fraction of image width/height
  (e.g. 0.04). 0.0 if clean.
\end{Verbatim}
\end{tcolorbox}

\subsection{Text OCR Prompt}
\label{app:ocr_prompt}
\begin{tcolorbox}[
    enhanced jigsaw,
    breakable,
    width=0.98\linewidth,
    colback=gray!5,
    colframe=gray!60,
    title=Prompt for Text OCR,
    fonttitle=\bfseries,
    pad at break*=1mm,
]
\scriptsize
\begin{Verbatim}[breaklines=true,breakanywhere=true]
You are an OCR transcription engine. Transcribe the text in this image
into one clean string, and output that string and NOTHING else.

ABSOLUTE OUTPUT RULES (follow exactly):
- Output ONLY the transcribed problem text. No preface, no notes, no commentary,
  no explanations, no markdown, no code fences.
- Do NOT think out loud or show any working. NEVER write meta words such as
  "Let's write", "Let's use", "Let me", "Wait", "the prompt says",
  "Correction", "Cleaning", or discuss how to format anything.
- NEVER list alternatives (e.g. 'X or X or X'). NEVER repeat a phrase. Write every
  piece of text exactly ONCE.
- Begin immediately with the first word of the problem. Stop immediately after the
  last option. Produce the transcription a single time, then stop.

TRANSCRIPTION RULES:
- Transcribe faithfully and completely. Do NOT solve, add, drop, paraphrase, or reorder.
- The image may wrap text mid-word or mid-line because of its width. Join it back into
  natural continuous prose and fix broken words (e.g. "th e right" -> "the right").
- Output as ONE continuous block with NO line breaks. Keep the multiple-choice options
  (A., B., C., D., ...) inline and in order, separated by a single space.
- Wrap mathematical content (symbols, variables, sub/superscripts, units inside a
  formula, equations) in single dollar signs, e.g. $v_0$, $R_2$, $2\text{ s}$,
  $a=\frac{3B^2d^2v_0}{2mR}$. Choose the simplest valid LaTeX and commit to it
  immediately; do not consider or show alternatives.
\end{Verbatim}
\end{tcolorbox}

\subsection{Visual Extractor Prompt}
\label{app:visual_extractor_prompt}
\begin{tcolorbox}[
    enhanced jigsaw,
    breakable,
    width=0.98\linewidth,
    colback=gray!5,
    colframe=gray!60,
    title=Prompt for Visual Extractor,
    fonttitle=\bfseries,
    pad at break*=1mm,
]
\scriptsize

\textbf{System Prompt}
\begin{Verbatim}[breaklines=true,breakanywhere=true]
You are a Visual Extractor for a physics problem-solving pipeline.
Convert the figure(s) into ONE self-contained SVG so a downstream
solver can reason about the geometry.
Output ONLY the raw <svg>...</svg> markup:
NO JSON, NO markdown fence, NO prose before or after.
\end{Verbatim}
\vspace{0.3em}
\textbf{User Prompt}
\begin{Verbatim}[breaklines=true,breakanywhere=true]
Render the figure(s) in this physics problem as a single SVG.

Problem text, possibly empty:
{problem}

SVG FORMAT:
- Use viewBox="0 0 100 100".
- Use ONLY line, circle, ellipse, rect, polygon, polyline, path, and text.
- Use semantic id attributes for important elements,
  e.g. id="mass_A", id="string", id="pulley",
  id="angle_30deg", id="field_B", id="label_25".
- Do NOT use <tspan>. Write any subscript/superscript
  INLINE as plain text inside the <text>.
- The SVG MUST end with </svg>.

CRITICAL EXTRACTION RULES:
- INCLUDE EVERY visible text label, number, variable,
  symbol, and unit shown in the figure as <text> elements.
- Every <text> element must correspond to visible text
  in the source figure.
- Transcribe all numeric text EXACTLY as shown,
  digit-for-digit, with its unit.
- Do NOT add, remove, or move decimal points.
- Do NOT invent missing values, velocities,
  directions, forces, constraints, or reference frames.
- Preserve the exact object-to-label association.

GEOMETRY RULES:
- Represent all visible objects, supports, pulleys,
  belts, strings, springs, coils, masses, blocks,
  tracks, inclines, rods, wires, axes, regions,
  fields, vectors, contact points, and angles.
- Preserve orientations and relative positions.
- Preserve topology exactly.
- Do NOT create or remove connections.
- Do NOT complete or infer hidden geometry.
- Draw labeled angles as arcs or paths and include
  the angle label exactly as shown.
  
STRICT FIDELITY RULES:
1. COPY NUMBERS EXACTLY.
2. TRANSCRIBE ONLY VISIBLE LABELS.
3. DO NOT OMIT ANY LABEL.
4. PRESERVE ANGLES AS SHOWN.
5. PRESERVE LABEL SCOPE.
6. PRESERVE TOPOLOGY.
7. HANDLE AMBIGUITY CONSERVATIVELY.

Before finalizing, silently verify:
- Every visible label appears as a <text> element.
- Every number matches the figure digit-for-digit.
- Every unit matches exactly.
- Every labeled angle is represented geometrically.
- Every visible object and connection is represented.
- No invisible or inferred physics has been added.
- The markup starts with <svg and ends with </svg>.
Output ONLY the raw SVG markup.
\end{Verbatim}
\end{tcolorbox}

\subsection{Visual-Grounded Problem Reconstructor Prompt}
\label{app:reconstructor_prompt}

\begin{tcolorbox}[
    enhanced,
    breakable,
    title=Prompt for Visual-Grounded Problem Reconstructor,
    colback=gray!5,
    colframe=gray!60,
    fonttitle=\bfseries,
    pad at break*=1mm,
]

\scriptsize

\textbf{System Prompt}

\begin{Verbatim}[breaklines=true,breakanywhere=true]
You convert a physics figure into a COMPLETE textual problem statement for a reader who CANNOT see any image.
\end{Verbatim}

\vspace{0.5em}

\textbf{User Prompt}

\begin{Verbatim}[breaklines=true,breakanywhere=true]
A physics problem is given with PARTIAL text; the rest of the information is ONLY in the attached figure.
Write the COMPLETE, self-contained problem statement in clear prose so someone with NO access to the figure can solve it.

CRITICAL ANTI-HALLUCINATION RULES:
- Transcribe ONLY what is explicitly written in the text or visibly shown in the figure.
- Do NOT infer, assume, deduce, or invent any physical condition, quantity, relationship, orientation, motion, or reference frame.
- Do NOT introduce descriptions such as 'relative to the belt', 'slides downward', 'moves upward',
  'hangs vertically', 'horizontal', 'vertical', or similar unless explicitly stated or visually indicated.
- If a value, label, direction, angle reference, or condition is ambiguous, illegible, or unclear,
  write 'ambiguous/unclear in the figure' rather than guessing.

REQUIREMENTS:
- Include EVERY numerical value, variable, symbol, label, unit, arrow, and annotation shown in the figure.
- Fully describe the physical setup and geometry:
  what is connected to what,
  object locations,
  orientations,
  reference axes,
  angle markings,
  angle references,
  initial conditions,
  and any indicated directions.
- Preserve the exact association between each quantity and the object it labels.
- Distinguish carefully between total quantities and per-object quantities.
- If multiple-choice, transcribe the EXACT wording of EVERY option (A, B, C, D, ...).
- Do NOT solve, simplify, interpret, explain, or add reasoning.
- Output ONLY the restated problem.

STRICT FIDELITY RULES (violating these corrupts the downstream solver):
1. TRANSCRIBE ONLY WHAT IS PRESENT.
   Do NOT add any value, variable, condition, relationship, or quantity that is not explicitly written or shown.
2. DO NOT ADD PHYSICAL RELATIONSHIPS.
   Do NOT introduce reference frames, motion descriptions, constraints, force relationships,
   equilibrium conditions, or geometric interpretations unless explicitly provided.
3. ATTRIBUTE QUANTITIES TO THE CORRECT OBJECT.
   Preserve exactly which object each number, label, mass, force, velocity, angle, length,
   charge, current, or other quantity belongs to.
   Do NOT redistribute, aggregate, or reinterpret quantities.
4. PRESERVE EVERY DIGIT AND UNIT EXACTLY.
   Copy all numbers character-for-character.
   Do NOT add, remove, or move decimal points.
   Do NOT change scale-looking numbers into decimals:
       25 must remain 25, not 2.5;
       15 must remain 15, not 1.5.
   Do NOT misread time labels:
       3 s must remain 3 s, not 35;
       7 s must remain 7 s, not 75.
   Examples:
   - '3 s' is NOT '35'
   - '7 s' is NOT '75'
   - '2.5' is NOT '25'
   - '0.5 m' is NOT '5 m'
5. PRESERVE ALL SYMBOLS AND NOTATION.
   Keep subscripts, superscripts, primes, vector arrows, Greek letters, signs,
   and mathematical notation exactly as shown.
6. PRESERVE EVERY ANGLE EXACTLY.
   Copy the angle value and its indicated reference.
   Do NOT convert an angle into an unstated orientation.
   For example, do NOT replace '30°' with 'vertical' or 'horizontal'
   unless explicitly indicated.
7. HANDLE AMBIGUITY CONSERVATIVELY.
   If any label, number, symbol, unit, angle reference, arrow direction,
   or condition is unclear, write
   'ambiguous/unclear in the figure'
   rather than guessing.

Before finalizing, silently verify:

- Every visible figure label appears in your output.
- No value appears that is not present in the source.
- Every number matches the source digit-for-digit.
- Every unit matches the source exactly.
- Every angle and its reference are preserved.
- No inferred physics has been added.

PARTIAL TEXT:
{ptext}

\end{Verbatim}
\end{tcolorbox}

\subsection{Round-1 Solver Prompt}
\label{app:solver_prompt}
\begin{tcolorbox}[enhanced jigsaw, breakable, width=0.98\linewidth, colback=gray!5,
    colframe=gray!60, title=Prompt for Round-1 Solver, fonttitle=\bfseries, pad at break*=1mm]
\scriptsize
\textbf{System Prompt}
\begin{Verbatim}[breaklines=true,breakanywhere=true]
You are an expert physics tutor.
CRITICAL ANSWER-FORMAT RULE: If the problem is MULTIPLE-CHOICE -- i.e. lettered
options (A, B, C, D, ...) appear in the text or the figure -- your final answer
MUST be ONLY the chosen option LETTER(S), e.g. 'C' or 'BD'. NEVER write the
option's words/value/concept (e.g. write 'C', not 'gravity').
For each problem, you must:
1. Carefully read the text and inspect any provided diagram(s).
2. Identify all known/unknown quantities and the physical principles involved.
3. Reason step-by-step with explicit equations.
4. Track units and report numeric answers to a sensible precision.
5. End with a clearly delimited final answer.
Output STRICTLY in this format:
<reasoning>
...your full chain of reasoning here...
</reasoning>
<answer>...the final answer only. If MULTIPLE-CHOICE, output ONLY the option
letter(s), e.g. A or BD -- NOT the option's text/content. Otherwise a concise
value (units only if asked), no extra text...</answer>
\end{Verbatim}
\vspace{0.3em}
\textbf{User Prompt}
\begin{Verbatim}[breaklines=true,breakanywhere=true]
Solve the following physics problem.

{problem}

Remember: produce exactly one <reasoning>...</reasoning> block followed by
exactly one <answer>...</answer> block. The final answer should be concise
(a number, expression, or option letter(s) like 'BD').
\end{Verbatim}
\end{tcolorbox}

\subsection{Debate (Re-solving) Prompt}
\label{app:debate_prompt}
\begin{tcolorbox}[enhanced jigsaw, breakable, width=0.98\linewidth, colback=gray!5,
    colframe=gray!60, title=Prompt for Re-solving (Debate), fonttitle=\bfseries, pad at break*=1mm]
\scriptsize
\textbf{System Prompt}
\begin{Verbatim}[breaklines=true,breakanywhere=true]
You are an expert physics tutor.
CRITICAL ANSWER-FORMAT RULE: If the problem is MULTIPLE-CHOICE (lettered options A,B,C,D... in the text or figure), the final answer MUST be ONLY the chosen option LETTER(S), e.g. 'C' or 'BD' — NEVER the option's words/value (write 'C', not 'gravity').
For each problem, you must:
1. Carefully read the text and inspect any provided diagram(s).
2. Identify all known/unknown quantities and the physical principles involved.
3. Reason step-by-step with explicit equations.
4. Track units and report numeric answers to a sensible precision.
5. End with a clearly delimited final answer.
6. If the problem is MULTIPLE-CHOICE (options shown in the text or figure), the final answer must be ONLY the option letter(s) (e.g. A, BD), NOT the option's text/content.
\end{Verbatim}
\vspace{0.3em}
\textbf{User Prompt}
\begin{Verbatim}[breaklines=true,breakanywhere=true]
These are the potential solutions to the problem:
{context}
Use the potential solutions as additional information for the following question.
An image of the figure is attached, and a structured parse of the SAME figure (PGDP) is included in the question below. As you reconsider, explicitly check whether any of the potential solutions (or your own previous reasoning) MISREAD or MISINTERPRETED the figure — e.g. wrong connectivity/topology, mislabeled or swapped points, incorrect geometry, or quantities/values read incorrectly from the diagram — and correct for it using the image and the PGDP parse.

Question:{question}
Please think step by step and solve the problem.

### Response format (MUST be strictly followed) (DO NOT include any other formats except for the given XML format): 
<think>YOUR THINKING HERE</think>
<answer>YOUR FINAL ANSWER ONLY, NO OTHER TEXT</answer>.
\end{Verbatim}
\end{tcolorbox}

\subsection{Pruning Prompt}
\label{app:prune_prompt}
\begin{tcolorbox}[enhanced jigsaw, breakable, width=0.98\linewidth, colback=gray!5,
    colframe=gray!60, title=Prompt for Pruning, fonttitle=\bfseries, pad at break*=1mm]
\scriptsize
\begin{Verbatim}[breaklines=true,breakanywhere=true]
Evaluate the given solution to the question. An image of the figure is attached, and a structured parse of the SAME figure (PGDP) is included in the Question below. ** Use BOTH the attached image and the PGDP parse. In particular, check whether the solution MISREAD or MISINTERPRETED the figure — e.g. wrong connectivity/topology, mislabeled or swapped points, incorrect geometry, or quantities/values read incorrectly from the diagram. If the solution's reasoning depends on any such misinterpretation of the figure, treat it as incorrect. ** ** Your reponse MUST end with the following format: <label>YES</label> or <label>NO</label> or <label>NOT SURE</label>. ** Return YES if the solution is completely correct, NO if any part of the solution is incorrect (including figure misinterpretation), and NOT SURE if you are unsure.

Question: {question}
Solutions: {solution}
\end{Verbatim}
\end{tcolorbox}

\subsection{Canonicalizer Prompt}
\label{app:canon_prompt}
\begin{tcolorbox}[enhanced jigsaw, breakable, width=0.98\linewidth, colback=gray!5,
    colframe=gray!60, title=Prompt for Canonicalizer, fonttitle=\bfseries, pad at break*=1mm]
\scriptsize
\begin{Verbatim}[breaklines=true,breakanywhere=true]
Render a physics FINAL ANSWER in clean grading-standard form. PRESERVE THE EXACT
VALUE. Include SI unit (infer if missing). Standard LaTeX (\frac, \sqrt,
\mathrm{}); no \boxed. Multiple-choice: ONLY letters. Output ONLY the answer.

\end{Verbatim}
\textbf{Tie-resolution Prompt (GPT-5.5; SYS\_GATE)}
\begin{Verbatim}[breaklines=true,breakanywhere=true]
You are a physics answer adjudicator. THREE solver answers (with reasoning) to ONE problem.
Compare FINAL ANSWERS by VALUE ONLY (ignore formatting/units/notation/labels/ordering/approx).
If >=2 share a value: RESOLVED: <that value, clean grading-standard form WITH SI unit; multiple-choice=ONLY letters; standard LaTeX, no \boxed>
If all 3 differ: ESCALATE
Output ONE line: 'RESOLVED: ...' or 'ESCALATE'.
\end{Verbatim}
\vspace{0.3em}
\textbf{Final-selection Prompt (GPT-5.5; SYS\_FINAL; used only on a three-way value tie after the last round)}
\begin{Verbatim}[breaklines=true,breakanywhere=true]
Three candidates, no value-majority. Pick the most correct final answer. Clean form, SI unit, multiple-choice=ONLY letters. Output ONLY the answer.
\end{Verbatim}
\end{tcolorbox}

%%%%%%%%%%%%%%%%%%%%%%%%%%%%%%%%%%%%%%%%%%%%%%%%%%%%%%%%%%%%%%%%%%%%%%%%%%%%%%%
%%%%%%%%%%%%%%%%%%%%%%%%%%%%%%%%%%%%%%%%%%%%%%%%%%%%%%%%%%%%%%%%%%%%%%%%%%%%%%%

\end{document}